# Distributed Consensus on Set–valued Information

Adriano Fagiolini, *Member, IEEE,* Nevio Dubbini, and Antonio Bicchi, *Fellow, IEEE*

*Abstract*—This paper focuses on the convergence of information in distributed systems of agents communicating over a network. The information on which the convergence is sought is not represented by real numbers, rather by *sets* of real numbers, whose possible dynamics are given by the class of so–called Boolean maps, involving only unions, intersections, and complements of sets. Based on a notion of contractivity, a necessary and sufficient condition ensuring the global and local convergence toward an equilibrium point is presented. In particular the analysis of global convergence recovers results already obtained by the authors, but the more general approach used in this paper allows analogue results to be found to characterize the local convergence.

## I. Introduction

Recent years have witnessed a constant migration of interests to applications involving many distributed agents that have to interact in order to achieve a common goal. Most of the problems, and of the solutions that have been proposed so far, can be formulated as consensus problems over *continuous domains*, where local agents exchange data that consists of real vectors or scalars. The only difference is in the type of rule each agent uses to combine its own information with the one received from its neighbors in the communication graph. In the simpler case, the evolution of the network of agents can be described by a linear iterative rule

$$x(t+1) = A\,x(t) + B\,u(t)\,,$$

where $t$ is a discrete time, $x \in \mathbb{R}^n$ is the system's state, $A$ is a weight matrix, and $u$ is an input vector. Matrix $A$ has to be compliant with the available communication graph and is designed to allow the network convergence to a unique decision $x(\infty) \to \alpha\,1$, that may or may not depend on the initial system's state. Falling into this linear framework are most of the key papers on consensus [1]–[3]. Moreover, the nonlinear setting encompasses other important schemes for achieving consensus on continuous, finite variables. Within this setting, the solution proposed by [4], based on the centroidal Voronoi tessellation, allows deployment of a collection of mobile agents so as to maximize the network's ability to perform a sensing task within a given environment. These problems, and indeed many others, can be efficiently solved by means of these agreement mechanisms.

However, new emerging issues in the field of distributed control entail defining different forms of consensus algorithms. Recently, [5] have addressed the sensing coverage problem with agents that are allowed to move in a discrete, network–like environment. The problem of averaging a set of initial measures, taken by a collection of distributed sensors, in the presence of communication constraints has recently been addressed by [6]. Therein, a consensus strategy, where exchanged data consists of symbols obtained through a logarithmic quantizer, is proposed. On the furthermost part of this track are problems over a discrete domain, where the system's state is a logical vector. This includes the problem of building a map of visitors/intruders' presence in the rooms and corridors of an art gallery, that has been attacked by [7], through introduction of so–called *logical consensus*.

Inadequacy of available solutions for distributed network agreement are apparent also in control applications where sensor measurement is affected by uncertainty. Consider e.g. a set of mobile robots that have to simultaneously localize themselves and build a map of the environment, by using their local vision systems. Traditional approaches to model sensors' noise as an additive or multiplicative signal is possible but not natural. [8] proposed a centralized solution, where robots exchange data representing confidence sets of the positions of items detected in the environment. Moreover, [9] considered the problem of detecting misbehaving agents within a collection of robots that are supposed to plan their motions according to a share set of rules. The objective is attained by definition of a *set–valued consensus algorithm*, where local agents exchange data representing free and occupied regions of the environment. The algorithm overcomes limitations of available solutions in the fact that it can operate over *infinite domains*. Finally, [10] and later [11] considered the problem of synchronizing the clocks of a set of distributed agents, and proposed a centralized solution to the problem. [12] have very recently shown that this problem can be solved by means of set–valued consensus.

So far, design of logical consensus as well as consensus on sets have been individually addressed, and only ad–hoc solutions have been proposed. As a matter of fact, a network of agents running either types of consensus are instances of a Boolean iterative system, i.e. a system where the state is a vector of elements in a Boolean domain and is updated through operations in a Boolean algebra. In [13] the authors present initial results toward the definition of a unified framework for dealing with such consensus problems. With this respect, a notion of a Boolean vector space is known since the seminal works of [14]–[16]. However, the behavior of a Boolean iterative system is far from been completely understood. The work is based on and extends some available results on the convergence of cellular automata [?], [17].

The main intent of the paper is to show that the convergence of information defined through discrete values or through set–valued data, combined via Boolean operations can be treated in the same way. This is achieved by extending the notions of convergence, local convergence, and contraction, already

Authors are with the Interdepartmental Research Center "E. Piaggio", Faculty of Engineering, Università di Pisa, Italy, {a.fagiolini, bicchi}@centropiaggio.unipi.it, dubbini@for.unipi.it.

given in in the binary domain, to algebras of sets, taken with the union, intersection, and complement operations. Then, we give results about global and local convergence of set–valued Boolean maps, through a binary encoding of them, which allows us to work only in the binary domain to solve these problems. Our main result is the characterization of the global and local convergence of set–valued maps obtained by means of unions, intersections, and complements, in terms of properties of binary matrices.

The paper is organized as follows. Section II recalls the definition of a Boolean Algebra (BA) and summarizes known results on the convergence of maps defined over the simplest BA, i.e. the ones involving a binary domain. These results are extended in the following sections. Section III studies the global behavior of Set–valued Boolean Maps (SBM). By using the binary encoding of a SBM that is presented in Section IV, conditions ensuring the global convergence of a SBM and local attractiveness of its equilibria are presented in Section V and Section VI, respectively. The problem of reaching consensus is solved for linear SBM in Section VII. Finally, examples of possible behaviors of SBM are presented in Section VIII.

## II. BOOLEAN DYNAMIC SYSTEMS

*Definition 1:* A Boolean Algebra (BA) is a sextuple $(\tilde{\mathbb{B}}, \wedge, \vee, \neg, 0, 1)$, consisting of a domain set $\tilde{\mathbb{B}}$, equipped with two binary operations $\wedge$ (called "meet" or "and") and $\vee$ (called "join" or "or"), a unary operation $\neg$ (called "complement" or "not"), and two elements 0 (null) and 1 (unity) belonging to $\tilde{\mathbb{B}}$, s.t. the following axioms hold, for all elements $a, b, c \in \tilde{\mathbb{B}}$:

1) $a \vee (b \vee c) = (a \vee b) \vee c$, $a \wedge (b \wedge c) = (a \wedge b) \wedge c$ (associativity);
2) $a \vee b = b \vee a$, $a \wedge b = b \wedge a$ (commutativity);
3) $a \vee (a \wedge b) = a$, $a \wedge (a \vee b) = a$ (absorption);
4) $a \vee (b \wedge c) = (a \vee b) \wedge (a \vee c)$, $a \wedge (b \vee c) = (a \wedge b) \vee (a \wedge c)$ (distributivity);
5) $a \vee \neg a = 1$, $a \wedge \neg a = 0$ (complementarity). ♦

From the first three pairs of axioms above, it follows that, for any two elements $a, b \in \tilde{\mathbb{B}}$, the following relation holds:

$$a = a \wedge b \quad \text{if, and only if,} \quad a \vee b = b,$$

which introduces a *partial order relation* $\leq$ among the elements of the domain. In particular, we will say that $a \leq b$, if, and only if, one of the two above equivalent conditions hold. Moreover, 0 and 1 are the least and greatest elements, respectively, of this partial order relation. Then, given any two elements $a, b \in \tilde{\mathbb{B}}$, the meet $a \wedge b$ and the join $a \vee b$ coincide with their infimum or supremum, respectively, w.r.t. $\leq$.

An element $a \in \tilde{\mathbb{B}}$ is referred to as a *scalar*. Consider the set $\tilde{\mathbb{B}}^n$ composed of Boolean *vectors* $x$, provided with the meet $\wedge$, and join $\vee$ with another vector $y \in \tilde{\mathbb{B}}^n$, and the meet $\wedge$ with a scalar $a \in \tilde{\mathbb{B}}$. Finally, consider the set of square Boolean matrices $A \in \tilde{\mathbb{B}}^{n \times n}$, provided with the meet and join operation between two matrices, and the meet of a matrix $A$ with a scalar $a$.

*Definition 2:* If $A = \{a_{ij}\}, B = \{b_{ij}\} \in \tilde{\mathbb{B}}^{n \times n}$ and $v = (v_1, \ldots, v_n)^T, w = (w_1, \ldots, w_n)^T \in \tilde{\mathbb{B}}^n$, we define the scalar product $w \cdot v$ to be

$$\bigvee_{i=1}^{n} v_i \wedge w_i \,,$$

$(A\,v)_i$ is defined as the scalar product between the $i$–th row of $A$ and the vector $v$, and $A\,B_{ij}$ to be the scalar product between the $i$-th row of $A$ and the $j$-th column of $B$.

In other words products between matrices and vectors, and between two matrices, are computed in the usual way, substituting $+$ with $\vee$ and $\cdot$ with $\wedge$. We will denote with 0 the null scalar, vector, or matrix, according to the context. The above described partial order relation $\leq$ between any two elements of $\tilde{\mathbb{B}}$ can be extended to Boolean vectors and matrices by assuming component–wise evaluation.

A Boolean dynamic system is s.t. its vector state, which takes value in $\tilde{\mathbb{B}}^n$, evolves according to a map $F$ that combines its input argument by using only the meet $\wedge$, join $\vee$, and complement $\neg$ operations.

For the following study, we need give the following definitions:

*Definition 3 (Basis Vectors):* The set of the vectors $e_1, e_2, \ldots, e_n$, with $e_j \in \tilde{\mathbb{B}}^n$ contains 1 in the $j$–th element and zeros elsewhere, is called the canonical basis of $\tilde{\mathbb{B}}^n$.

*Definition 4 (Eigenvalues and Eigenvectors):* A scalar $\lambda \in \tilde{\mathbb{B}}$ is an *eigenvalue* of a Boolean matrix $A \in \tilde{\mathbb{B}}^{n \times n}$ if there exists a vector $x \in \tilde{\mathbb{B}}^n$, called *eigenvector*, s.t.

$$A\,x = \lambda\,x \,.$$

*Definition 5 (Incidence Matrix):* The incidence matrix of a Boolean map $F$ is a Boolean binary matrix $B(F) = \{b_{i,j}\}$, where $b_{i,j} = 1$ if, and only if, the $i$–th component of $F(x)$ depends on the $j$–th component of the input vector $x$.

### A. Convergence of Binary Dynamic Systems

Consider the simplest BA that is obtained by choosing $\tilde{\mathbb{B}} = \mathbb{B}$, where $\mathbb{B} = \{0, 1\}$ is the binary domain, the meet $\wedge$ is the logical product (and) $\cdot$ (corresponding to the minimum of the two input arguments), the join $\vee$ is the logical sum (or) $+$ (corresponding to the maximum of the two input arguments), the complement $\neg$ is the not operator, 0 and 1 are the binary values. Consider the following notion:

*Definition 6 (Binary Spectral Radius):* The spectral radius of a Boolean matrix $A \in \tilde{\mathbb{B}}^{n \times n}$, denoted with $\rho(A)$, is its biggest eigenvalue in the sense of the vector order relation $\leq$. Consider the following propositions from [?], [17]:

*Proposition 1:* Every binary matrix $A \in \mathbb{B}^{n \times n}$ has at least one eigenvalue. Hence $\rho(A)$ always exists.

*Proposition 2:* A binary matrix $A \in \mathbb{B}^{n \times n}$ has binary spectral radius $\rho(A) = 0$ if, and only if, one of the two following equivalent conditions hold:
- there exists a permutation matrix $P \in \mathbb{B}^{n \times n}$ s.t. $P^T A\,P$ is a strictly lower or upper triangular matrix;
- $A^n = 0$ (meaning the $n$–th binary matrix power of $A$).

Consider studying the evolutions of a dynamic system of the form

$$\begin{cases} x(t+1) = f(x(t)) \,, \\ x(0) = x^0 \,, \end{cases}$$

where $x = (x_1, \ldots, x_n)^T \in \mathbb{B}^n$ is the system's binary state, $f : \mathbb{B}^n \to \mathbb{B}^n$, $f = (f_1^T, \cdots, f_n^T)^T$, is an endomorphism on $\mathbb{B}^n$, and $x^0$ is the system's initial state. As $\mathbb{B}^n$ is a finite set, the convergence of the state sequence generated by iterations of $f$ corresponds to the fact that the sequence itself becomes stationary after a certain time $\bar{t}$. To characterize this convergence, we first need to introduce a metric on $\mathbb{B}^n$. To this aim, consider the binary vector distance

$$\begin{aligned} d \; : \; & \mathbb{B}^n \times \mathbb{B}^n \to \mathbb{B}^n \\ & (x,y) \mapsto (x_1 \oplus y_1, \cdots, x_n \oplus y_n) \end{aligned},$$

where $\oplus$ is the exclusive disjunction

$$\begin{aligned} \oplus \; : \; & \mathbb{B} \times \mathbb{B} \to \mathbb{B} \\ & (x_i, y_i) \mapsto (\neg x_i \, y_i) + (x_i \, \neg y_i) \end{aligned},$$

The binary vector distance is a distance on $\mathbb{B}^n$ as, for all $x, y, z \in \mathbb{B}^n$, it satisfies the axioms

$$\begin{aligned} d(x,y) &= d(y,x), \\ d(x,y) &= 0 \text{ iff } x = y, \\ d(x,y) &\leq d(x,z) + d(z,y). \end{aligned}$$

Consider also the following notion:

*Definition 7 (Contractive Map):* A map $f : \mathbb{B}^n \to \mathbb{B}^n$ is said to be contractive w.r.t. the binary vector distance $d$ if there exists a matrix $M \in \mathbb{B}^{n \times n}$ s.t.

- $\rho(M) < 1$ (which implies $\rho(M) = 0$), and
- $d(f(x), f(y)) \leq M \, d(x,y)$, for all vectors $x, y \in \mathbb{B}^n$.

We can readily recall the main result on the contractiveness of a binary map [?]:

*Theorem 1:* A map $f : \mathbb{B}^n \to \mathbb{B}^n$ is contractive w.r.t. the binary vector distance $d$ if, and only if, the following equivalent conditions hold:

- $\rho(B(f)) = 0$;
- there exists a permutation matrix $P$ s.t. $P^T B(f) P$ is strictly lower or upper triangular;
- $B(f)^q = 0$, with $0 \leq q \leq n$.

Moreover, if $f$ is contractive, there exists a positive integer $q \leq n$ s.t. $f^q$, the composition of $f$ with itself $q$ times, is a constant map, i.e. it does not depend on the input vector. ♦

Now let us focus on the local convergence [17] of a map $f$ about an *equilibrium point* $x$ s.t. $f(x) = x$:

*Definition 8 (Von–Neumann Neighborhood (VNN)):* Given a point $x \in \mathbb{B}^n$, its VNN is the set $V(x)$ of all points differing from $x$ in at most one component, i.e.

$$V(x) = \{x, \tilde{x}^1, \cdots, \tilde{x}^n\},$$

where $\tilde{x}^j = (x_1, \cdots, x_{j-1}, \neg x_j, x_{j+1}, \cdots, x_n)^T$.

If $\mathbb{B}^n$ is represented as a hypercube and all its elements as its vertices, $\tilde{x}^j$ can be interpreted as the $j$-th vertex adjacent to $x$. Note that, for all $x \in \mathbb{B}^n$, $d(x, \tilde{x}^j) = e_j$ and $d(x, 0) = x$.

*Definition 9 (Discrete Derivative):* The discrete derivative of a binary map $f : \mathbb{B}^n \to \mathbb{B}^n$ at a generic point $x \in \mathbb{B}^n$ is a binary matrix $f'(x) = \{f'_{i,j}\}$, s.t. $f'_{i,j} = 1$ if, and only if, a variation in the $j$-th component of $x$ produces a variation in the $i$-th component of $f(x)$, i.e.,

$$f'_{i,j}(x) = f_i(x) \oplus f_i(\tilde{x}^j).$$

It is worth noting that, the assignment of a vector $x \in \mathbb{B}^n$, the value of the logical map $f(x)$, the value of its discrete derivative $f'(x)$ at that point, uniquely determines the value $f(y)$ of the logical map at any point $y$ in the immediate neighborhood $V(x)$. Furthermore, if $f'(x) = 0$, then $f(y)$ is constant and equal to $f(x)$, for all points $y \in V(x)$.

Finally, consider the following two notions:

*Definition 10:* An equilibrium point $x \in \mathbb{B}^n$ is said to be *attractive* in its VNN $V(x)$ if the following two relations hold:

- $f(y) \in V(x)$, for all $y \in V(x)$;
- there exists $n \in \mathbb{N}$ s.t., for all $y \in V(x)$, $f^n(y) = x$.

*Definition 11:* A binary map $f$ is said to be *locally convergent* at an equilibrium point $x$ if $x$ is attractive in its VNN.

We can recall the main result on the attractiveness of an equilibrium point [17]:

*Theorem 2:* An equilibrium point $x \in \mathbb{B}^n$ is attractive in its VNN if, and only if, the following two relations hold:

- $\rho(f'(x)) = 0$,
- $f'(x)$ contains at most one non–null element in each column. ♦

*Example 2.1:* Consider the map

$$f(x) = \begin{pmatrix} x_3(x_1 + \neg x_2) \\ x_3(x_1 + x_2) + \neg x_3(\neg x_1 + x_2) \\ x_1 \end{pmatrix}.$$

Its incidence matrix is

$$B(f) = \begin{pmatrix} 1 & 1 & 1 \\ 1 & 1 & 1 \\ 1 & 0 & 0 \end{pmatrix},$$

whose spectral radius is $\rho(B(f)) = 1$, which tells us that the map is not contractive. Thus, the presence of multiple equilibria or cycles cannot be excluded. Indeed, let us find the equilibria states $\bar{x}$ being s.t. $f(\bar{x}) = \bar{x}$, or equivalently

$$\neg(f_i(\bar{x}) \oplus \bar{x}_i) = 1 \text{ for } i = 1, 2, 3.$$

After some simplification, this gives the binary equations $\bar{x}_1 \bar{x}_3 + \neg \bar{x}_1 \neg \bar{x}_3 = 1$, $\bar{x}_1 \bar{x}_3 + \neg \bar{x}_1 \neg \bar{x}_2 + \neg \bar{x}_2 \bar{x}_3 + \neg \bar{x}_1 \neg \bar{x}_3 = 1$, and $\neg \bar{x}_1 (\bar{x}_2 + \bar{x}_3) + \bar{x}_2 + \bar{x}_3 = 1$, which are solved by the vectors

$$\begin{aligned} \bar{x}^{(1)} &= (0, 1, 0) \\ \bar{x}^{(2)} &= (1, 1, 1). \end{aligned}$$

Moreover, the discrete derivatives of the map $f$ at the two equilibria are

$$f'\left(\bar{x}^{(1)}\right) = \begin{pmatrix} 0 & 0 & 0 \\ 0 & 0 & 0 \\ 1 & 0 & 0 \end{pmatrix}, \; f'\left(\bar{x}^{(2)}\right) = \begin{pmatrix} 1 & 0 & 1 \\ 0 & 0 & 0 \\ 1 & 0 & 0 \end{pmatrix},$$

which tell us, based on Theorem 2, that the first equilibrium in $\bar{x}^{(1)}$ is attractive in its VNN, whereas the second one in $\bar{x}^{(2)}$ is not.

To conclude consider Fig. 2.1 that is a graphical representation of the binary map $f$, which can be obtained by complete inspection of the map itself. The figure clearly shows that the VNN of $\bar{x}^{(1)}$ "contracts" toward it, whereas this is not true for $\bar{x}^{(2)}$. The figure also shows the existence of a cycle composed of the states $(0,0,1)$ and $(1,0,0)$. ♦

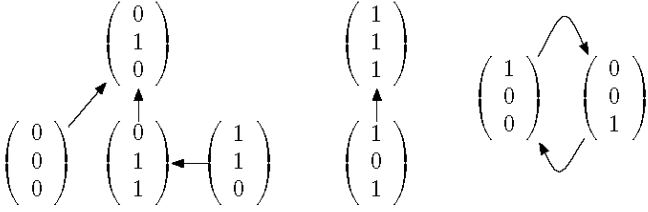

Figure 1. Graphical representation of the binary map $f$ of Example 2.1.

## III. Set–valued Boolean Dynamic Systems - Global Convergence

Thanks to Stone's Representation Theorem [18], it is well known that every BA is isomorphic (i.e. it possesses the same structural properties) to an algebra of sets. Let us then focus on the BA defined by the sextuple $(\mathcal{P}(\mathbb{X}), \cup, \cap, \mathcal{C}(\cdot), \emptyset, \mathbb{X})$, where $\mathcal{P}(\mathbb{X})$ is the power set of a set $\mathbb{X}$, and on dynamic systems of the form

$$X(t+1) = F(X(t)), \qquad (1)$$

where $X = (X_1, \cdots, X_n)^T$, with $X_i \in \mathcal{P}(\mathbb{X})$, is a set–valued state vector, and $F = (F_1, \cdots, F_n)^T$, with $F_i : \mathcal{P}(\mathbb{X})^n \to \mathcal{P}(\mathbb{X})$, is a set–valued map involving only the operations $\cup$, $\cap$, $\mathcal{C}(\cdot)$ on continuous and possibly unbounded sets. Such a class of maps will be referred to as Set–valued Boolean Maps (SBM). In other words, the map $F$ combines $n$ input sets into $n$ output sets via unions, intersections, and complements. In the remainder of this section, we study under which conditions these systems converge to a unique equilibrium.

First, we need to introduce a metric over the Boolean vector space $\mathcal{P}(\mathbb{X})^n$. To this aim, consider the *Boolean vector distance*:

$$\begin{aligned} \mathcal{D} \ : \ & \mathcal{P}(\mathbb{X})^n \times \mathcal{P}(\mathbb{X})^n \to \mathcal{P}(\mathbb{X})^n \\ & (X,Y) \mapsto (S(X_1,Y_1), \cdots, S(X_n,Y_n)) \end{aligned}$$

where $X_i, Y_i$ are the $i$–th components of the vectors $X$ and $Y$, respectively, and $S$ is the symmetric difference defined as

$$\begin{aligned} S \ : \ & \mathcal{P}(\mathbb{X}) \times \mathcal{P}(\mathbb{X}) \to \mathcal{P}(\mathbb{X}) \\ & (x,y) \mapsto (\mathcal{C}(x) \cap y) \cup (x \cap \mathcal{C}(y)) \end{aligned}$$

Note that the Boolean vector distance $\mathcal{D}$ specializes to the binary vector distance of [?], when considering the binary BA, and it satisfies the following axioms

$$\mathcal{D}(X,Y) = \mathcal{D}(Y,X), \ \forall \ X, Y,$$
$$\mathcal{D}(X,Y) = \emptyset \text{ iff } X = Y,$$
$$\mathcal{D}(X,Y) \subseteq \mathcal{D}(X,Z) \cup \mathcal{D}(Z,Y).$$

Consider the following notion:

*Definition 12 (Incidence matrix):* The incidence matrix of a set–valued map $F = (F_1, \cdots, F_n)^T$, with $F_i : \mathcal{P}(\mathbb{X})^n \to \mathcal{P}(\mathbb{X})$, denote with $B(F)$, is a binary Boolean matrix whose generic element is

$$b_{i,j} = \begin{cases} \mathbb{X} & \text{if } F_i \text{ depends on } X_j, \\ \emptyset & \text{otherwise}. \end{cases}$$

We can recall the following results concerning the incidence matrix of a Boolean map [13]:

*Proposition 3:* Given any two generic vectors $X, Y \in \mathcal{P}(\mathbb{X})^n$, the following Boolean inequality holds

$$\mathcal{D}(F(X), F(Y)) \subseteq B(F)\, \mathcal{D}(X,Y). \blacklozenge \qquad (2)$$

*Proposition 4:* A Boolean matrix $M$ satisfies the Boolean inequality

$$\mathcal{D}(F(X), F(Y)) \subseteq M \mathcal{D}(X,Y), \qquad (3)$$

for all vectors $X, Y \in \mathcal{P}(\mathbb{X})^n$, if, and only if, $B(F) \subseteq M$. $\blacklozenge$

*Corollary 1:* For any two set–valued maps $F, G : \mathcal{P}(\mathbb{X})^n \to \mathcal{P}(\mathbb{X})^n$, the incidence matrix of the function composition $F(G(X))$ satisfies the Boolean inequality

$$B(F(G)) \subseteq B(F)\, B(G). \blacklozenge \qquad (4)$$

Moreover, consider the following notion:

*Definition 13 (Boolean spectrum):* The Boolean spectrum of a Boolean matrix $A \in \mathcal{P}(\mathbb{X})^{n \times n}$ is set of the eigenvalues of $A$, i.e.,

$$\sigma(A) = \{ \lambda \in \mathcal{P}(\mathbb{X}) \mid \exists\, x \in \mathcal{P}(\mathbb{X}) \setminus \emptyset \,:\, A\, x = \lambda\, x \}. \blacklozenge$$

We can recall from [13] a first result about the spectrum of a Boolean map:

*Proposition 5:* A Boolean matrix $A \in \mathcal{P}(\mathbb{X})^{n \times n}$, $A = \{a_{i,j}\}$, admits the Boolean eigenvalue $\lambda = \emptyset$ if, and only if, it has at least one column for which the union of all its elements is less than $\mathbb{X}$, i.e. there exists $j \in \{1, \cdots, n\}$ s.t.

$$\bigcup_{i=1}^{n} a_{i,j} \subset \mathbb{X}. \blacklozenge \qquad (5)$$

*Remark 1:* It is worth noting that, if $A$ has a Boolean eigenvalue $\lambda$, with assigned eigenvector $X$, then, for every permutation $P$, the matrix $A' = P^T A P$ has the same eigenvalue, assigned with eigenvector $V = P^T X$. Note that $P$ is a permutation in the classical sense, but where every 0 and 1 are replaced with $\emptyset$ and $\mathbb{X}$, respectively.

To prove this, observe that $A X = \lambda X$ for hypothesis. Left–multiplying by $P^T$, we have $P^T A X = \lambda\, P^T X$, and, from the identity $P^T P = I$ ($I$ being the matrix with $X$ on its diagonal and $\emptyset$ elsewhere), we have $(P^T A P)(P^T X) = \lambda\, (P^T X)$, which proves the statement. $\blacklozenge$

A complete characterization of the Boolean spectrum of a generic map is complex, whereas the following result is already available for a subclass of these maps:

*Proposition 6:* A matrix $A \in \{\emptyset, \mathbb{X}\}^{n \times n}$ admits the Boolean eigenvalue $\lambda = \mathbb{X}$ if, and only if, there exist no permutation bringing $A$ in strictly lower or upper triangular form. $\blacklozenge$

With the following example, we show that the spectrum of a Boolean matrix may possess a structure that is impossible in $\mathbb{R}^n$, e.g. different eigenvalues can be associated with the same eigenvector, or the spectrum may be the entire set $\mathcal{P}(\mathbb{X})$.

*Example 3.1:* Consider the entire real interval $\mathbb{X} = (-\infty, \infty)$. The matrix

$$A_1 = \begin{pmatrix} \emptyset & \emptyset \\ (17, 28] & 13 \end{pmatrix}$$

admits the eigenvalue $\lambda = \emptyset$, by Prop. 5, as the union of its first row's elements is less than $\mathbb{X}$. The associated eigenvectors are $V_\lambda = (X, \emptyset)^T$, where $X$ is any set in $\mathbb{X}$.

Moreover, consider the matrix

$$A_2 = \begin{pmatrix} [3,5) & \mathbb{X} \\ \mathbb{X} & 4 \end{pmatrix}.$$

By direct computation, it can be shown that any scalar $\lambda \subseteq \mathbb{X} \setminus \emptyset$ is an eigenvalue $A_2$, with associated eigenvector $V_\lambda = (X, X)^T$, with $\lambda \subseteq X$. ♦

Moreover, consider the following definition:

*Definition 14 (Contractive SBM):* A SBM $F : \mathcal{P}(\mathbb{X})^n \to \mathcal{P}(\mathbb{X})^n$ is said to be *contractive* w.r.t. the vector distance if

$$\mathbb{X} \notin \sigma(B(F)).$$

*Remark 2:* From Prop. 6, $F$ is contractive if there exists a permutation matrix $P \in \{\emptyset, \mathbb{X}\}^{n \times n}$ s.t. $P^T B(F) P$ is strictly lower or upper triangular. ♦

Finally, recall from [13] the following two results characterizing the global convergence of a SBM $F$:

*Theorem 3:* $F$ is contractive w.r.t. the vector distance $\mathcal{D}$ if, and only if, there exists a positive integer $q$ s.t. $F^q$ is a constant application. ♦

*Corollary 2:* If $\xi$ is the unique equilibrium point, iterations of $F$ starting from any initial state $X(0) \in \mathcal{P}(\mathbb{X})^n$ converge to $\xi$ in at most $q$ steps. ♦

*Example 3.2:* Consider a discrete–time dynamic system $X(t+1) = F(X(t))$, where $X = (X_1, X_2, X_3)^T \in \mathcal{P}(\mathbb{X})^3$ and $F$ is the SBM

$$F(X) = \begin{pmatrix} X_1 \cup (X_2 \cap X_3) \\ X_1 \cup \mathcal{C}(X_2) \\ \mathcal{C}(X_1) \cap \mathcal{C}(X_2) \cap \mathcal{C}(X_3) \end{pmatrix}. \quad (6)$$

Its incidence matrix is

$$B(F) = \begin{pmatrix} \mathbb{X} & \mathbb{X} & \mathbb{X} \\ \mathbb{X} & \mathbb{X} & \emptyset \\ \mathbb{X} & \mathbb{X} & \mathbb{X} \end{pmatrix},$$

and, based on Theorem 3, its spectrum contains the eigenvalue $\lambda = \mathbb{X}$, which tells us that the map is not contractive.

## IV. BINARY ENCODING OF SET–VALUED BOOLEAN DYNAMIC SYSTEMS

In this section, we show how a dynamic system of the form in Eq. 1 can be translated into a binary dynamic system

$$x(t+1) = f(x(t)), \quad (7)$$

where $x \in \mathbb{B}^\kappa$ is binary state vector, and $f : \mathbb{B}^\kappa \to \mathbb{B}^\kappa$ is a binary iterative map, and $\kappa$ is a suitable dimension [19]. We say that the system in Eq. 7 encodes the original system in Eq. 1, in the sense that every execution of the original system can be obtained by simulating the binary one and vice–versa

Consider the collection of sets

$$\begin{aligned} Z_1 &= X_1 \cap X_2 \cap \cdots \cap X_{n-1} \cap X_n, \\ Z_2 &= X_1 \cap X_2 \cap \cdots \cap X_{n-1} \cap \mathcal{C}(X_n), \\ Z_3 &= X_1 \cap X_2 \cap \cdots \cap \mathcal{C}(X_{n-1}) \cap X_n, \\ &\vdots \\ Z_{\kappa'-1} &= \mathcal{C}(X_1) \cap \mathcal{C}(X_2) \cap \cdots \cap \mathcal{C}(X_{n-1}) \cap X_n, \\ Z_{\kappa'} &= \mathcal{C}(X_1) \cap \mathcal{C}(X_2) \cap \cdots \cap \mathcal{C}(X_{n-1}) \cap \mathcal{C}(X_n), \end{aligned}$$

with $\kappa' = 2^n$. Let us denote with $Z = (Z_1, Z_2, \ldots, Z_\kappa)^T$ the vector composed of the non–empty sets of the previous collection (note that in general $\kappa \leq \kappa'$). It is straightforward to verify that these sets are a partition of $\mathbb{X}$, i.e. $X_i \cap X_j = \emptyset$, and $X_1 \cup \cdots \cup X_n = \mathbb{X}$. In the remainder of this section, we show that every set $X_i \subseteq \mathbb{X}$, and indeed all unions, intersections, and complements obtained from the $X_i$ can be obtained as the union of some of the above sets, which allows us to find a binary encoding of the SBM $F$.

Consider an *encoder map* associating a set $X_i$ with a binary vector whose $h$–th component is 1 if, and only if, $X_i$ has non–null overlapping with the set $Z_h$, i.e.

$$\begin{aligned} \mathcal{L} &: \mathcal{P}(\mathbb{X}) \to \mathbb{B}^\kappa \\ X_i &\mapsto x_i = \begin{pmatrix} x_1^i \\ \vdots \\ x_\kappa^i \end{pmatrix}, \quad x_h^i = \begin{cases} 0 & \text{if } X_i \cap Z_h = \emptyset, \\ 1 & \text{otherwise}. \end{cases} \end{aligned}$$

Thus, given a set $X_i$, the corresponding associated binary vector is $x_i = \mathcal{L}(X_i)$. The original set $X_i$ can be obtained via the *decoder map*

$$\begin{aligned} \mathcal{L}^{-1} &: \mathbb{B}^\kappa \to \mathcal{P}(\mathbb{X}) \\ x_i &\mapsto X_i = \bigcup\nolimits_{h=1,\cdots,\kappa,\, x_h^i = 1} Z_h, \end{aligned}$$

which allows us to write $X_i = \mathcal{L}^{-1}(x_i)$. Furthermore, consider any two sets, $X_i$ and $X_j$, of the given collection, and their logical encoded vectors, $x_i = \mathcal{L}(X_i)$, and $x_j = \mathcal{L}(X_j)$. Consider first the their combination via set intersection:

$$\begin{aligned} X_i \cap X_j &= \mathcal{L}^{-1}(x_i) \cap \mathcal{L}^{-1}(x_j) = \\ &= \left( \bigcup\nolimits_{h=1,\, x_h^i = 1}^\kappa Z_h \right) \cap \left( \bigcup\nolimits_{l=1,\, x_l^j = 1}^\kappa Z_l \right), \end{aligned}$$

which can be expanded, by distributing the set intersection, as the union of the sets given by the intersection of one $Z_h$ with one $Z_l$. As all $Z_i$ are disjoint, only those $Z_i$ appearing in both the original sets, $X_i$ and $X_j$, remain in the intersection. Therefore, we can write

$$\begin{aligned} X_i \cap X_j &= \bigcup\nolimits_{h=1,\, (x_h^i = 1) \wedge (x_h^j = 1)}^\kappa Z_h = \\ &= \bigcup\nolimits_{h=1,\, x^h = 1}^\kappa Z_h = \mathcal{L}^{-1}(x), \text{ with } x = x_i x_j, \end{aligned}$$

which proves the equivalence relation:

$$X_i \cap X_j \underset{\mathcal{L}^{-1}}{\overset{\mathcal{L}}{\rightleftharpoons}} x_i x_j. \quad (8)$$

Moreover, consider the two sets' combination via set union:

$$\begin{aligned} X_i \cup X_j &= \mathcal{L}^{-1}(x_i) \cup \mathcal{L}^{-1}(x_j) = \\ &= \left( \bigcup\nolimits_{h=1,\, x_h^i = 1}^\kappa Z_h \right) \cup \left( \bigcup\nolimits_{l=1,\, x_l^j = 1}^\kappa Z_l \right) = \\ &= \bigcup\nolimits_{h=1,(x_h^i = 1 \vee x_h^j = 1)}^\kappa Z_h = \bigcup\nolimits_{h=1,\, x^h = 1}^\kappa Z_h = \\ &= \mathcal{L}^{-1}(x), \text{ with } x = x_i + x_j, \end{aligned}$$

which proves the equivalence relation:

$$X_i \cup X_j \underset{\mathcal{L}^{-1}}{\overset{\mathcal{L}}{\rightleftharpoons}} x_i + x_j. \quad (9)$$

Finally, consider the complementation of one of the two sets:

$$\begin{aligned} \mathcal{C}(X_i) &= \mathcal{C}(\mathcal{L}^{-1}(x_i)) = \mathcal{C}\left(\bigcup_{h=1,\,x_h^i=1}^{\kappa} Z_h\right) = \\ &= \bigcap_{h=1,\,x_h^i=1}^{\kappa} \mathcal{C}(Z_h)\,. \end{aligned}$$

By definition $\mathcal{C}(Z_h)$ is the set of points not belonging to $Z_h$, that can be obtained as the union of all the other partition sets:

$$\begin{aligned} \bar{Z}_h &= \mathcal{C}(Z_h) = \bigcup_{h'=1,h'\neq h}^{\kappa} Z'_h = \\ &= Z_1 \cup Z_2 \cup \cdots \cup Z_{h-1} \cup Z_{h+1} \cup \cdots \cup Z_{\kappa} = \\ &= \mathcal{L}^{-1}(z_1) \cup \mathcal{L}^{-1}(z_2) \cup \cdots \cup \mathcal{L}^{-1}(z_{h-1}) \cup \\ &\quad \cup \mathcal{L}^{-1}(z_{h+1}) \cup \cdots \cup \mathcal{L}^{-1}(z_\kappa) = \\ &= \bigcup_{l=1,\alpha_{l,h}=1}^{\kappa} Z_l\,, \end{aligned}$$

with $\alpha_h = z_1 + z_2 + \cdots + z_{h-1} + z_{h+1} + \cdots + z_\kappa$. Easy computation gives a logical vector $\alpha_h = (1,\ldots,1,0,1,\ldots,1)^T$ containing all entries to 1 except for the $h$-th one. Finally, intersection of all $\bar{Z}_h$ yields:

$$\begin{aligned} \mathcal{C}(X_i) &= \bigcap_{h=1,\,x_h^i=1}^{\kappa} \bar{Z}_h = \\ &= \bigcap_{h=1,\,\alpha^h=1}^{\kappa} Z_h,\text{ with }\alpha = \alpha_1\alpha_2\ldots\alpha_r\,, \end{aligned}$$

where $r$ is the number of $x_i$'s non–null components. As all these components are assigned with a logical vector $\alpha_l$ containing a null element at position $l$, and as all these components are considered, the sets that remain in the intersection are those not belonging to $X_i$, or in other words, for which $x_h^i = 0$. Hence, we have

$$\begin{aligned} \mathcal{C}(X_i) &= \bigcup_{h=1,\,x_h^i=0}^{\kappa} Z_h = \bigcup_{h=1,\,x_h^i=1}^{\kappa} Z_h = \\ &= \mathcal{L}^{-1}(y_i),\text{ with }y_i = \neg x_i\,, \end{aligned}$$

which proves the equivalence relation:

$$\mathcal{C}(X_i) \underset{\mathcal{L}^{-1}}{\overset{\mathcal{L}}{\rightleftharpoons}} \neg x_i\,. \tag{10}$$

*Remark 3:* From the above results, it follows that the intersection (union, complement) of any two sets $X_i$ and $X_j$ is equivalent, via the encoder map, to the bitwise logical product (sum, complement) of the corresponding binary vectors $x_i = \mathcal{L}(X_i)$ and $x_j = \mathcal{L}(X_j)$.

In Boolean logic, the notion of Algebraic Normal Form (ANF) [20] of a binary map is often used for formal theorem proving. The following proposition gives a generalization of this notion for SBMs:

*Proposition 7 (Normal Form (NF)):* For each SBM $F : \mathcal{P}(\mathbb{X}) \to \mathcal{P}(\mathbb{X})$ depending on $n$ sets, the Normal Form

$$F(X) = \mathcal{D}_{\substack{k=0,\cdots,n \\ J \in S(n,k)}} \left(A_J \cap \left(\bigcap_{j \in J} X_j\right)\right)\,,$$

where $A_J \in \{\emptyset, \mathbb{X}\}$, $X_\emptyset = \mathbb{X}$, and $S : \mathbb{N} \times \mathbb{N} \to 2^{2^\mathbb{N}}$ is s.t. $S(n,k)$ are all combinations of $k$ elements out of the first $n$ integers in lexicographical order ($S(n,0) = \{\{\emptyset\}\}$ by definition), fully describes $F$. Moreover, it holds

$$A_J = \mathcal{D}\left(F(y_J), \mathcal{D}_{\substack{k=0,\cdots,n_J \\ I \in S(J,k)}} (A_I)\right)\,,$$

where $n_J = \text{card}(J) - 1$ and $y_J = (y_1, \cdots, y_n)^T$, with $y_i = \mathbb{X}$

| $F(X_1)$ | $A_\emptyset$ | $A_{\{1\}}$ |
|---|---|---|
| $\emptyset, X_1 \cap \emptyset, X_1 \cap \neg X_1$ | $\emptyset$ | $\emptyset$ |
| $\mathbb{X}, X_1 \cup \neg X_1, X_1 \cup \mathbb{X}$ | $\mathbb{X}$ | $\emptyset$ |
| $X_1, X_1 \cup X_1, X_1 \cap X_1, X_1 \cup \emptyset, X_1 \cap \mathbb{X}$ | $\emptyset$ | $\mathbb{X}$ |
| $\mathcal{C}(X_1)$ | $\mathbb{X}$ | $\mathbb{X}$ |

Table I
SUMMARY OF NF'S SET COEFFICIENTS FOR A ONE PARAMETER SET–VALUED BOOLEAN FUNCTION.

if $i \in J$, or $\bar{x}_i = \emptyset$ otherwise.

*Proof:* Let us proceed by induction. The NF of a SBM depending on $n = 1$ input set is

$$\begin{aligned} F(X) &= \mathcal{D}(A_\emptyset, A_{\{1\}} \cap X_1) = \\ &= \left(A_\emptyset \cap \mathcal{C}(A_{\{1\}})\right) \cup \left(\mathcal{C}(A_\emptyset) \cap A_{\{1\}} \cap X_1\right) \cup \\ &\quad \cup \left(A_\emptyset \cap \mathcal{C}(X_1)\right)\,. \end{aligned}$$

To show that the proposition holds it is sufficient to show that, for each function of one input argument, there exists a choice for the sets $A_\emptyset$ and $A_{\{1\}}$ correctly describing the function (see Table 7). ■

As an example of Prop. 7, consider a map computing the union of its two input sets, i.e., $F(X_1, X_2) = X_1 \cup X_2$. The NF of a generic map with $n = 2$ arguments is

$$\begin{aligned} F(X_1, X_2) &= A_\emptyset \cup \\ &\quad \cup \left(A_{\{1\}} \cap X_1\right) \cup \left(A_{\{2\}} \cap X_2\right) \cup \\ &\quad \cup \left(A_{\{1,2\}} \cap X_1 \cap X_2\right)\,, \end{aligned}$$

with $A_\emptyset, A_{\{1\}}, A_{\{2\}}, A_{\{1,2\}} = \{\emptyset, \mathbb{X}\}$. Set union can be obtained by choosing $A_\emptyset = \emptyset$, $A_{\{1\}}, A_{\{2\}}, A_{\{1,2\}} = \mathbb{X}$, i.e. $F(X_1, X_2) = X_1 \cup X_2 \cup (X_1 \cap X_2)$. Moreover, in the binary case, saying that a binary map is fully described by its Algebraic Normal Form (ANF) means that the two maps have the same truth table.

We can now state the main result of this section, which proves that Remark 3 is valid for any SBM:

*Theorem 4 (Binary Encoding of SBM):* A dynamic system of the form

$$X(t+1) = F(X(t))\,,$$

where $F$ is a generic SBM, with initial state $X(0)$, can be simulated by $\kappa$ copies of a binary dynamic system

$$x^i(t+1) = f(x^i(t))\,,\quad\text{for } i = 1, \cdots, \kappa\,,$$

where the binary map is

$$f : \mathbb{B}^n \to \mathbb{B}$$
$$\begin{pmatrix} x_1 \\ \vdots \\ x_n \end{pmatrix} \mapsto d_{k=1}^n \left(d_{J \in S(n,k)} a_J \prod_{j \in J} x_j\right)\,,$$

in which $\kappa$ is determined by the binary encoding, $a_J = \mathcal{L}(A_J)$ for all $J$, $A_J$ are the coefficient sets of the NF of $F$, and initial states given by $x = \mathcal{L}(X(0))$.

*Proof:* First note that every $A_J$ can be associated with logical vectors via the above encoding:

$$a_J = \mathcal{L}(A_J) = \begin{cases} (0, \cdots, 0)^T & \text{if } A_J = \emptyset, \\ (1, \cdots, 1)^T & \text{if } A_J = \mathbb{X}. \end{cases}$$

The NF of a generic SBM $F(X)$ can be rewritten as

$$F(X) = \mathcal{D}_{\substack{k=0,\cdots,n \\ J \in S(n,k)}} Y_J,$$

where $Y_J = A_J \cap \left(\bigcap_{j \in J} X_j\right)$. Note that every set $Y_J$ can be associated with a logical vector $y_J = \mathcal{L}(Y_J)$ as it is given by the intersection of sets that can be associated with logical vectors by the above encoding (see the equivalence relation in Eq. 8). Therefore, we can write

$$F(X) = \mathcal{D}_{\substack{k=0,\cdots,n \\ J \in S(n,k)}} \mathcal{L}^{-1}(y_J),$$

with $y_J = \mathcal{L}(Y_J) = a_J \prod_{j \in J} x_j$.

Furthermore, the distance between two vector sets is $\mathcal{D}(X,Y) = (\mathcal{C}(X) \cap Y)) \cup (X \cap \mathcal{C}(Y))$, which involves only unions, intersections, and complements. Based on the equivalence relations in Eq. 9, 8, 10, the set–valued distance $\mathcal{D}$ is equivalent to the binary distance $d$. Therefore, we can also write $F(X) = \mathcal{L}^{-1}(d_{k=0}^n \alpha_k)$, with $\alpha_k = d_{I \in S(n,k)} y_J$. This implies that $F(X)$ can be evaluated by decoding the result of the logical function

$$f(x) = d_{k=1}^n \left( d_{J \in S(n,k)} a_J \prod_{i \in J} x_i \right),$$

where $a_J = \mathcal{L}(A_J)$ for all $J$, which gives the thesis. ∎

*Corollary 3:* $f$ is obtained from $F$ by $k$ copies of a function obtained by replacing every set $X_i \in \mathcal{P}(\mathbb{X})$ with a binary vector $x_i = \mathcal{L}(X_i) \in \mathbb{B}^\kappa$ and all unions, intersections, and complements with logical bitwise sums, products, and negations, respectively. In this sense, we will write

$$f = \mathcal{L}(F).$$

*Proof:* Consider the sequence of operations reducing the NF of $F$ into its original expression. A similar sequence can be applied to the ANF of $f$ of Theorem 4, where each operation in the original algebra is replaced with the corresponding in the binary ones. ∎

*Example 4.1:* Consider again the system of Example 3.2. We want to compute the corresponding binary dynamic system. The partition sets are

$$\begin{aligned}
Z_1 &= X_1 \cap X_2 \cap X_3, \\
Z_2 &= X_1 \cap X_2 \cap \mathcal{C}(X_3), \\
Z_3 &= X_1 \cap \mathcal{C}(X_2) \cap X_3, \\
Z_4 &= X_1 \cap \mathcal{C}(X_2) \cap \mathcal{C}(X_3), \\
Z_5 &= \mathcal{C}(X_1) \cap X_2 \cap X_3, \\
Z_6 &= \mathcal{C}(X_1) \cap X_2 \cap \mathcal{C}(X_3), \\
Z_7 &= \mathcal{C}(X_1) \cap \mathcal{C}(X_2) \cap X_3, \\
Z_8 &= \mathcal{C}(X_1) \cap \mathcal{C}(X_2) \cap \mathcal{C}(X_3),
\end{aligned}$$

Each state $X_i$ is associated with a binary vector $x_i \in \mathbb{B}^8$. Based on Theorem 4, the corresponding logical system is $x(t+1) = f(x(t))$, with $x = (x_1^T, x_2^T, x_3^T)^T$ and

$$\begin{aligned}
f(x) = \; &(x_{1,1} + x_{2,1}x_{3,1}, \cdots, x_{1,8} + x_{2,8}x_{3,8}, \\
&x_{1,8}\bar{x}_{2,8}, \cdots, x_{1,8}\bar{x}_{2,8}, \\
&\bar{x}_{1,1}\bar{x}_{2,1}\bar{x}_{3,1}, \cdots, \bar{x}_{1,8}\bar{x}_{2,8}\bar{x}_{3,8}).
\end{aligned}$$

Consider a numeric example in which the unity is $\mathbb{X} = [0, \infty)$ and the original system is initialized with the sets $X_1(0) = [2,5]$, $X_2(0) = [4,7]$, and $X_3(0) = [8,11]$. According to the update rule in Eq. 6, the original system's state after one step is

$$X(1) = F(X(0)) = \begin{pmatrix} [2,5] \cup ([8,11] \cap [4,7]) \\ [2,5] \cup \mathcal{C}([8,11]) \\ \mathcal{C}([2,5]) \cap \mathcal{C}([8,11]) \cap \mathcal{C}([4,7]) \end{pmatrix} =$$

$$= \begin{pmatrix} [2,5] \cup \emptyset \\ [2,5] \cup [0,8) \cup (11, \infty) \\ ([0,2) \cup (5, \infty)) \cap A \end{pmatrix},$$

with $A = ([0,8) \cup (11, \infty)) \cap ([0,4) \cap (7, \infty))$, which yields

$$X(1) = \begin{pmatrix} [2,5] \\ [0,5] \cup (7, \infty) \\ [0,2) \cup (7,8) \cup (11, \infty) \end{pmatrix}. \quad (11)$$

The same result can be obtained via the binary encoding. The values of the partition sets are:

$$Z_1 = [4,5], Z_2 = [2,4), Z_3 = [5,7],$$
$$Z_4 = [8,11], Z_5 = [0,2) \cup (7,8) \cup (11, \infty).$$

The initial state of the corresponding binary dynamic system is is $x(0) = (x_1^T(0), x_2^T(0), x_3(0)^T)^T)^T$, with $x_i(0) = \mathcal{L}(X_i(0))$, for $i = 1,2,3$, and is given by

$$x_1(0) = (1,1,0,0,0), \; x_2(0) = (1,0,1,0,0),$$
$$x_3(0) = (0,0,0,1,0).$$

Application of the logical iteration $f$ gives the next state $x(1) = f(x(0))$ with

$$x_1(1) = (1,1,0,0,0), \; x_2(1) = (1,1,0,1,1),$$
$$x_3(1) = (0,0,0,0,1).$$

This corresponds to

$$X(1) = \mathcal{L}^{-1}(x(1)) = \begin{pmatrix} Z_1 \cup Z_2 \\ Z_1 \cup Z_2 \cup Z_4 \cup Z_5 \\ Z_5 \end{pmatrix},$$

which gives the same result as in Eq. 11.

## V. GLOBAL CONVERGENCE OF SET–VALUED BOOLEAN SYSTEMS - A REVISED VIEW

We now present the first main result of the paper, concerning the way of proving that a Boolean map is contractive (i.e. converge) globally. Suppose that a map $F : \mathcal{P}(\mathbb{X})^n \to \mathcal{P}(\mathbb{X})^n$ is given, s.t. it can be described using unions, intersections, and complements only. Let $f : \mathbb{B}^{n \times \kappa} \to \mathbb{B}^{n \times \kappa}$ be its translation $\mathcal{L}(F)$. By Theorem 3, $F$ is contractive $\Leftrightarrow$ if there exists a positive integer $q$ s.t. $F^q$ is a constant application $\Leftrightarrow$ $\mathbb{X} \notin \sigma(B(F))$. We show that properties such as the map's contractivity can indeed be investigated in the binary domain.

*Lemma 1:* Define $B(f)$ as the incidence matrix of $f$ in the same way as $B(F)$ is for $F$. Then

$$\{B(F)\}_{i,j} = \mathbb{X}$$
$$\Leftrightarrow$$
$$\{B(f)\}_{2^\kappa(i-1)+1:2^\kappa(i-1), 2^\kappa(j-1)+1:2^\kappa(j-1)} = I.$$

where $I$ is the identity matrix.

*Proof:* Write the map $F$ in the following way:

$$F(X_1,\ldots,X_n) = \begin{pmatrix} F_1\left(X_{i_1^1},\ldots,X_{i_{k_1}^1}\right) \\ \vdots \\ F_n\left(X_{i_1^n},\ldots,X_{i_{k_n}^n}\right) \end{pmatrix},$$

where the $X_{i_l^j}$ are the actual variables on which the $l$-th component of the image of $F$ depends. Now, from the definition of the map $\mathcal{L}$, we have

$$B(f) = \begin{pmatrix} 0 & \cdots & 0 & \overbrace{Id}^{i_1^1} & 0 & \cdots & \overbrace{Id}^{i_2^1} \\ \vdots & \cdots & \cdots & \cdots & \cdots & \cdots & \cdots \\ 0 & \cdots & \underbrace{Id}_{i_1^n} & 0 & \cdots & \underbrace{Id}_{i_2^n} & 0 \\ & & & \overbrace{Id}^{i_{k_1}^1} & \cdots & 0 & \\ \cdots & & & \cdots & \cdots & \vdots & \\ \cdots & & & \underbrace{Id}_{i_{k_n}^n} & \cdots & 0 & \end{pmatrix},$$

$(B(f) \in \mathbb{B}^{n\kappa \times n\kappa})$ where every 0 and $I$ are zero matrices and identity matrices, respectively. The thesis now follows easily since $B(F) \in \{\emptyset, \mathbb{X}\}^{n\times n}$ has exactly the same form, substituting $\emptyset$ to 0 and $\mathbb{X}$ to $I$:

$$B(F) = \begin{pmatrix} \emptyset & \cdots & \emptyset & \overbrace{\mathbb{X}}^{i_1^1} & \emptyset & \cdots & \overbrace{\mathbb{X}}^{i_2^1} \\ \vdots & \cdots & \cdots & \cdots & \cdots & \cdots & \cdots \\ \emptyset & \cdots & \underbrace{\mathbb{X}}_{i_1^n} & \emptyset & \cdots & \underbrace{\mathbb{X}}_{i_2^n} & \emptyset \\ & & & \overbrace{\mathbb{X}}^{i_{k_1}^1} & \cdots & \emptyset & \\ \cdots & & & \cdots & \cdots & \vdots & \\ \cdots & & & \underbrace{\mathbb{X}}_{i_{k_n}^n} & \cdots & \emptyset & \end{pmatrix}.$$

■

*Theorem 5:* Suppose that $F : \mathcal{P}(\mathbb{X})^n \to \mathcal{P}(\mathbb{X})^n$ uses unions, intersections, and complements only. Then $F$ is contractive if, and only if, $\mathcal{L}(F) : \{0,1\}^{n\kappa} \to \mathbb{B}^{n\kappa}$ is contractive.

*Proof:* By Theorems 1 and 3 it is sufficient to prove that

$$\mathbb{X} \notin \sigma(B(F)) \Leftrightarrow \rho(B(f)) = 0.$$

By remark 2 $\mathbb{X} \notin \sigma(B(F))$ if, and only if, there exists a permutation matrix $P$ s.t. $P^T B(F) P$ is strictly lower or upper triangular, while proposition 2 assures that $\rho(B(f)) = 0$ is and only if there exists a permutation matrix $p$ s.t. $p^T B(f) p$ is strictly lower triangular. Now, by Lemma 1 it holds

$$\{B(F)\}_{ij} = \mathbb{X}$$
$$\Leftrightarrow$$
$$\{B(f)\}_{2^{n(i-1)}+1:2^{n(i-1)}, 2^{n(j-1)}+1:2^{n(j-1)}} = I.$$

This immediately implies that

$$\mathbb{X} \notin \sigma(B(F)) \Leftrightarrow \rho(B(f)) = 0.$$

■

*Remark 4:* $\rho(B(f)) = 0$ if and only if $\rho\left(\tilde{B}(F)\right) = 0$, where $\tilde{B}(f)$ is the matrix obtained substituting 1 to $\mathbb{X}$ and 0 to $\emptyset$. This can be easily seen using the equivalent formulations in terms of permutation matrices given by proposition 2 and remark 2. ◆

## VI. Local Convergence of Set–valued Boolean Dynamic Systems

In the same way we can investigate the local convergence properties of a set-valued map in the binary domain: while equivalent formulations in boolean domain exist for global convergence, this method give a novel technique to prove local convergence (see also Remark 6).

*Definition 15:* Given a generic vector $X = (X_1,\ldots,X_j,\ldots,X_n)^T \in \mathcal{P}(\mathbb{X})^n$, its $j$-th *neighbor* is

$$\tilde{X}^j = (X_1,\ldots,\mathcal{C}(X_j),\ldots,X_n)^T. \blacklozenge$$

*Definition 16 (Neighborhood):* Given a vector $X \in \mathcal{P}(\mathbb{X})^n$, neighborhood of $X$ is the set $V(X)$ of all points that differ in at most one complemented component from $X$:

$$V(X) = \{X, \tilde{X}^1, \cdots, \tilde{X}^n\}. \blacklozenge$$

*Definition 17:* An equilibrium point $X$ of $F : \mathcal{P}(\mathbb{X})^n \to \mathcal{P}(\mathbb{X})^n$ is *attractive* in its neighborhood if
- $F(V(X)) \subset V(X)$;
- $F^n(Y) = X \quad \forall Y \in V(X)$.

Suppose now given a map $F : \mathcal{P}(\mathbb{X})^n \to \mathcal{P}(\mathbb{X})^n$, that can be described only in terms of unions, intersections, and complements. Computing $\mathcal{L}(F)$ we obtain the map $f : \mathbb{B}^{n\cdot\kappa} \to \mathbb{B}^{n\cdot\kappa}$.

*Theorem 6:* A map $F : \mathcal{P}(\mathbb{X}^n) \to \mathcal{P}(\mathbb{X}^n)$ that can be described only in terms of unions, intersections, and complements, is attractive in the immediate neighborhood of an equilibrium set $X$ if and only if $\mathcal{L}(F) : \mathbb{B}^{n\cdot\kappa} \to \mathbb{B}^{n\cdot\kappa}$ is attractive in the immediate neighborhood of the equilibrium point $x$ corresponding to $X$.

*Proof:* The key point of the proof is the fact that, once given an initial condition, we can indifferently see the dynamics of $F$ in $\mathcal{P}(\mathbb{X})^n$ or the dynamics of $f$ in $\mathbb{B}^{n\kappa}$. In particular it's straightforward to verify comparing definitions 8 and 16, that

$$F(V(X)) \in V(X) \Leftrightarrow f(V(x)) \in V(x),$$
$$F^n(Y) = X, \ \forall Y \in V(X) \Leftrightarrow f^n(y) = x, \ \forall y \in V(x).$$

Now Theorem 2 implies the thesis. ■

*Remark 5:* In other words Theorem 5 states that we can check local attractivity by working on binary matrices.

*Remark 6:* It's worth noting that the proof of theorem 2 *cannot* be adapted to the local convergence in set-algebra domain. Indeed the notion of discrete derivative has its natural set-domain generalization in the definition of boolean derivative given in [19]. This definition however gives rise to set-matrices whose elements are not only the empty and the whole set: this prevent from using the techniques in [13]. In this paper there is no need of the notion of a derivative in the set domain. ♦

In the following proposition we show that Theorems 5 and 6 can be applied when the definition of the map $F$ involves other (usual) operations, that can be defined in terms of unions, intersections, and complements, also with the help of some tricks.

*Proposition 8:* Theorems 5 and 6 can be applied also when the definition of the map $F$ involves set difference, symmetric difference, or constant sets.

*Proof:* Set difference is defined to be
$$X \setminus Y = \{x \in X \,:\, x \notin Y\} = \{X \cap \mathcal{C}(Y)\},$$
which proves that the set difference can be defined only with unions, intersections, and complements. Symmetric difference is defined to be $\mathcal{D}(X,Y) = X \cup Y \setminus X \cap Y$, which proves that, also symmetric difference can be defined only with unions, intersections, and complements. ∎

With regards to the maps involving not only operations among the arguments, but also operations involving constant sets, say $A_1, \ldots, A_k$, we use the following trick: define another function $\tilde{F}$, modifying adding $k$ new components in the definition of $F$

$$F\begin{pmatrix} X_1 \\ \vdots \\ X_n \\ X_{n+1} \\ \vdots \\ X_{n+k} \end{pmatrix} = \begin{pmatrix} F(X_1, \ldots, X_n, \mathcal{A}_1, \ldots, \mathcal{A}_k) \\ \vdots \\ F(X_1, \ldots, X_n, \mathcal{A}_1, \ldots, \mathcal{A}_k) \\ \mathcal{A}_1 \\ \vdots \\ \mathcal{A}_k \end{pmatrix}.$$

In this way the map $\tilde{F}$ contains only unions, intersections, and complements among its arguments, and Theorems 5, 6 apply. ♦

*Remark 7:* Thanks to proposition 8, given a map $F : \mathcal{P}(\mathbb{X}^n) \to \mathcal{P}(\mathbb{X}^n)$, defined in terms of intersections, unions, complements, set differences, symmetric differences, and involving also constant sets, Theorems 5, 6 apply: we can check contractivity and local convergence working only with binary matrices.

## VII. APPLICATION TO LINEAR SBM

Consider a linear system $X^+ = AX$ of the form
$$\begin{pmatrix} X_1^+ \\ \vdots \\ X_n^+ \end{pmatrix} = \begin{pmatrix} a_{11} & \ldots & a_{1n} \\ \vdots & \ddots & \vdots \\ a_{n1} & \ldots & a_{nn} \end{pmatrix} \begin{pmatrix} X_1 \\ \vdots \\ X_n \end{pmatrix} = \begin{pmatrix} a_{11} \cap X_1 \cup \ldots \cup a_{1n} \cap X_n \\ \vdots \\ a_{n1} \cap X_1 \cup \ldots \cup a_{nn} \cap X_n \end{pmatrix}.$$

Following proposition 8, we have to consider the following system, given by the introduction of matrix constants as additional variables which remains constant in time:

$$\begin{cases} X_1^+ = a_{11} \cap X_1 \cup \ldots \cup a_{1n} \cap X_n \\ \vdots \\ X_n^+ = a_{n1} \cap X_1 \cup \ldots \cup a_{nn} \cap X_n \\ X_{n+1}^+ = a_{11} \\ \vdots \\ X_{n+n^2}^+ = a_{nn} \end{cases}$$

By remark 4, we can write the incidence matrix with 1 instead of $\mathbb{X}$, and without the map $\mathcal{L}$:

$$\begin{pmatrix} \begin{pmatrix} \ulcorner B(a_{11}) & \ldots & B(a_{1n}) \urcorner \\ \vdots & \ddots & \vdots \\ \llcorner B(a_{n1}) & \ldots & B(a_{nn}) \lrcorner \end{pmatrix} & \begin{matrix} \ulcorner 0 & \ldots & 0 \urcorner \\ 0 & \ldots & 0 \\ \vdots & \ddots & \vdots \\ \llcorner 0 & \ldots & 0 \lrcorner \end{matrix} \\ \begin{matrix} \ulcorner B(a_{11}) & \ldots & B(a_{1n}) \urcorner \\ \vdots & \ddots & \vdots \\ \llcorner 0 & \ldots & 0 \lrcorner \end{matrix} & \begin{matrix} \ulcorner 0 & \ldots & 0 \urcorner \\ \vdots & \ddots & \vdots \\ \llcorner B(a_{n1}) & \ldots & B(a_{nn}) \lrcorner \end{matrix} \\ & \mathbf{Id}_{n^2 \times n^2} \end{pmatrix},$$

where $B(a_{ij}) = 1$ if $a_{ij} \neq 0$, and $B(a_{ij}) = 0$ otherwise. Now, global contractivity is equivalent, by Theorem 3, to the fact that $A^q$ is a constant application. This happens if and only if the matrix $B(A) = \begin{pmatrix} B(a_{11}) & \cdots & B(a_{1n}) \\ \vdots & \ddots & \ddots \\ B(a_{n1}) & \cdots & B(a_{nn}) \end{pmatrix}$ is nilpotent, and this latter condition holds, by remark 2, if and only if $\rho(B(A)) = 0$, that is if there exists a permutation matrix $P$ s.t. $P^T B(A) P$ is strictly triangular.

*Proposition 9 (Consensus of linear systems):* Simple calculations shows that a linear systems of type $X^+ = AX$, has a consensus fixed point if and only if
$$\bigcap_{i=1}^n a_{i1} \cup a_{i2} \cup \ldots \cup a_{in} \neq \emptyset.$$

*Proof:* Observe that there exists a consensus fixed point $\Phi$ if and only if

$$\begin{pmatrix} \Phi \cap (a_{11} \cup a_{12} \cup \ldots \cup a_{1n}) \\ \vdots \\ \Phi \cap (a_{n1} \cup a_{12} \cup \ldots \cup a_{nn}) \end{pmatrix} = \begin{pmatrix} \Phi \\ \vdots \\ \Phi \end{pmatrix}.$$

This implies that the matrix $A$ has a consensus algorithm if and only if the intersection in the statement of the proposition

Figure 2. Iterations of the system (12) given by initial conditions (a). The first and the second iteration are shown in (b) and (c) respectively, while the third iteration, (d) is the fixed point of the system.

is non-empty. ♦

## VIII. EXAMPLES

*Example 8.1:* Consider the following map $F : \mathcal{P}([0, 200]) \rightarrow \mathcal{P}([0, 200])$ given by the following updating rule:

$$\begin{aligned} x_1^+ &= x_3 \cup (x_2 \cap x_5) \\ x_2^+ &= x_3 \\ x_3^+ &= x_3(0) \\ x_4^+ &= x_1 \cup (x_2 \cap x_3) \cup (x_5 \cap x_6) \\ x_5^+ &= x_2 \cap x_3 \\ x_6^+ &= (x_1 \cap x_3) \cup x_2 \cup x_5 \end{aligned} \quad (12)$$

The incidence matrix $B(F)$ is then

$$B(F) = \begin{pmatrix} 0 & 1 & 1 & 0 & 1 & 0 \\ 0 & 0 & 1 & 0 & 0 & 0 \\ 0 & 0 & 0 & 0 & 0 & 0 \\ 1 & 1 & 1 & 0 & 1 & 1 \\ 0 & 1 & 1 & 0 & 0 & 0 \\ 1 & 1 & 1 & 0 & 1 & 0 \end{pmatrix}.$$

The permutation matrix $P = \begin{pmatrix} 0 & 0 & 0 & 1 & 0 & 0 \\ 0 & 1 & 0 & 0 & 0 & 0 \\ 1 & 0 & 0 & 0 & 0 & 0 \\ 0 & 0 & 0 & 0 & 0 & 1 \\ 0 & 0 & 1 & 0 & 0 & 0 \\ 0 & 0 & 0 & 0 & 1 & 0 \end{pmatrix}$ is s.t. $P^T B(F) P$ is a strictly lower triangular matrix, therefore the system (12) is contractive, by Theorems 5 and 1. In figure 8.1 a particular orbit is shown, which converge to a consensus given by the initial condition $x_3(0)$. Moreover, since the system is contractive, the fixed point does not depends on the initial conditions, and the system always reaches a consensus given by the initial condition $x_3(0)$. ♦

## IX. CONCLUSIONS

This paper focused on the convergence towards consensus on information in distributed systems, where agents share data that is not represented by real numbers, rather by logical values or sets. We showed that both types of information convergence problems can indeed be attacked in a unified way in the framework of Boolean distributed information systems. Based on a notions of contractivity and local convergence for Boolean dynamical systems, a necessary and sufficient condition ensuring the global and local convergence toward an equilibrium point is presented. Application of achieved results to some examples was finally shown. Future works will address the convergence of more general set-valued maps.